\documentclass[runningheads]{llncs}

\usepackage{eccv}

\usepackage{eccvabbrv}

\usepackage{graphicx}
\usepackage{booktabs}
\usepackage{tabularx}
\usepackage[ruled]{algorithm2e}
\usepackage[accsupp]{axessibility}  % Improves PDF readability for those with disabilities.

\usepackage{hyperref}

\usepackage{orcidlink}
\usepackage{multirow}
\usepackage{enumitem}
\begin{document}

\title{Confidence-Aware Teacher-Student Distillation for 3D Medical Segmentation} 


\author{Georgios Triantafyllou \orcidlink{0009-0001-2807-9211} \and
Dimitris K. Iakovidis \thanks{Corresponding author.}\orcidlink{0000-0002-5027-5323}}

\authorrunning{G. Triantafyllou, D. K. Iakovidis}

\institute{Department of Computer Science and Biomedical Informatics, University of Thessaly, 2-4 Papasiopoulou st. 35131, Lamia, Greece \\
\email{\{gtriantafyllou,diakovidis\}@uth.gr}}

\maketitle

\begin{abstract}
Medical image segmentation models typically rely on large amounts of densely annotated volumetric data, limiting their scalability across tasks and imaging modalities. This work addresses the challenge of predicting entire 3D anatomical structures from extreme annotation sparsity. An annotation-efficient student–teacher framework is proposed for automatic 3D medical segmentation that requires only a set of point prompts on a single 2D slice per volume, as input. A foundation model serves as an offline teacher, utilizing the provided point prompts from the selected slice to full-volume pseudo-annotations alongside their corresponding spatial confidence scores prior to student training. To mitigate the error propagation of noisy pseudo-annotations, a task-specific 3D student network is trained using a confidence-aware optimization strategy. By leveraging the teacher's pre-computed confidence scores, this strategy explicitly excludes statically uncertain regions of the pseudo-annotations from the loss calculation, while simultaneously emphasizing regions with higher confidence. Evaluated on 3D cardiac MRI datasets, our framework outperforms state-of-the-art semi-supervised methods, improving segmentation performance by up to 43.60\%. Furthermore, it drastically reduces the manual annotation burden to just a few positive point prompts per volume, while improving surface boundary precision by up to 14.70\% over the teacher and successfully recovering up to 34.10\% of the performance gap toward the fully supervised upper bound.
\end{abstract}

\section{Introduction}
\label{sec:intro}

Medical image segmentation underpins diagnosis, treatment planning, and quantitative biomarker extraction \cite{marinov2024deep}. While deep neural networks have substantially advanced segmentation accuracy, most high-performing systems remain computationally expensive and rely heavily on dense expert annotation. This requirement for dense annotation remains a major bottleneck for scaling medical image segmentation across different organs, scanners, and imaging modalities \cite{feng2024enhancing}. Even when public benchmarks exist, annotations are frequently limited to a small subset of anatomies or acquisition protocols, and clinical translation often requires adapting to new do-mains where expert delineation is prohibitively costly.
\par
To reduce reliance on dense annotations, semi-supervised learning exploits limitedly annotated data together with data that are not annotated, commonly through pseudo-annotations, consistency regularization, and knowledge-based priors \cite{jiao2024learning}. Additionally, large pretrained segmentation foundation models, including the Segment Anything Model (SAM) \cite{kirillov2023segment} and its medical adaptations \cite{ma2024segment}, can serve as external supervision sources by generating plausible pseudo-annotations for task-specific networks using sparse cues (e.g., scribbles or point prompts). However, significant practical challenges remain. First, prompt-based foundation models require continuous per-image user interaction to generate accurate image segmentation, rendering them inefficient for direct deployment \cite{shen2024fastsam3d}. Second, the generated pseudo-annotations are often noisy and unreliable, particularly near anatomical boundaries or in low-contrast regions \cite{hang2025pseudo}. While recent frameworks, such as  SemiSAM+ \cite{zhang2025semisam+} attempt to mitigate this noise through online collaborative loops, they still dynamically query a foundation model during student training. Consequently, this online coupling introduces severe computational and memory overheads during optimization. 
\par

To address these limitations, an annotation-efficient, confidence-aware student–teacher framework is introduced for 3D medical image segmentation. The approach is designed for extreme annotation sparsity, e.g., even if only one weakly annotated image, using point prompts, is available. A frozen pretrained foundation model is utilized as an offline teacher to generate initial pseudo-annotations, \textit{i.e.}, pseudo-label segmentation masks, and per-pixel confidence estimates based on the input point prompts. To limit the propagation of errors contained in these pseudo-labels, a 3D student network is subsequently trained using a confidence-aware objective that excludes low-confidence voxels from optimization and weights the retained supervision according to confidence. This optimization strategy explicitly emphasizes highly reliable regions while discarding ambiguous predictions.
\par
To this end, the main contributions of this study are:

\begingroup
\setlength{\emergencystretch}{1em}

\begin{itemize}[
    label=\textbullet,
    labelindent=2.5em, % indentation of the bullet from the left
    leftmargin=*,      % automatically calculate text indentation
    labelsep=0.6em,
    topsep=0.1em,
    itemsep=0.3em,
    parsep=0pt,
    partopsep=0pt
]
    \item An annotation-efficient training pipeline that distills pseudo-annotations generated from a pre-trained foundation model into a 3D student network, requiring only a small set of point prompts on a single 2D slice per volumetric scan.

    \item A confidence-aware optimization strategy that materializes reliable pseudo-label targets and masks out uncertain regions prior to training, effectively mitigating pseudo-annotation noise.

    \item A student-teacher decoupling strategy at inference time that retains the high-accuracy benefits of foundation-model priors while fully discarding the teacher network at deployment, ensuring efficient execution.

    \item Experimental validation on two different 3D cardiac MRI datasets, demonstrating that the proposed framework achieves improved 3D segmentation accuracy over relevant state-of-the-art approaches, and recovers a substantial portion of the performance gap toward fully supervised models despite the extreme reduction in annotation cost.
\end{itemize}

\endgroup

\section{Related Work}

To alleviate the annotation bottleneck, semi-supervised learning (SSL) methods exploit unlabeled data through consistency regularization and pseudo-label supervision. The Mean Teacher approach \cite{tarvainen2017mean} enforced consistency between a student net-work and an exponential moving average (EMA) teacher. Building on this paradigm, SGRS-Net \cite{wang2025synergy} partitioned teacher-generated pseudo-labels into regions and applied region-specific supervision, while AD-MT \cite{zhao2024alternate} alternated between two momentum teachers to provide diverse supervisory signals. Furthermore, mutual supervision methods provided a complementary approach. Specifically, Cross Pseudo Supervision (CPS) \cite{chen2021semi} exchanged pseudo-labels between parallel networks, with subsequent medical segmentation methods incorporating high-confidence guidance \cite{cheng2025semi} or combining fixed and dynamically generated pseudo-labels \cite{kumari2025leveraging}. MC-Net+ \cite{wu2022mutual} adopted a multi-predictor formulation, using a shared encoder and heterogeneous decoders whose outputs are regularized through mutual consistency in ambiguous regions. In addition, FixMatch-based \cite{sohn2020fixmatch} approaches employed confidence-aware self-training under weak and strong augmentations, while recent work has also investigated uncertainty-aware consistency and contrastive learning \cite{assefa2025dycon}. Despite their different formulations, most of these methods derive at least part of their supervision from predictions updated during optimization. Erroneous targets may therefore be reinforced during training, particularly near anatomical boundaries and within low-contrast regions, thereby exacerbating confirmation bias \cite{hang2025pseudo}. 

\par
The emergence of large pretrained foundation models, such as SAM \cite{kirillov2023segment} and its medical adaptations (e.g., MedSAM \cite{ma2024segment}), has demonstrated strong zero-shot generalization capabilities for 2D images. Building upon this, next-generation architectures, such as SAM2 \cite{ravi2024sam} and its domain-specific variants, such as MedSAM2 \cite{ma2025medsam2} have recently extended these capabilities to video and 3D volumetric data. By treating 3D medical scans as sequential frames, these advanced models can propagate sparse cues, such as point prompts provided on a single spatial slice, across the entire volume to generate continuous 3D segmentation masks. Despite their versatility and the significant reduction in manual annotation effort, utilizing these foundation models directly in clinical workflows presents practical challenges. While they alleviate the need for exhaustive per-slice interaction, they still fundamentally rely on iterative user feedback to correct and refine the resulting segmentation masks. Furthermore, their continuous memory mechanisms and heavy parameter count introduce substantial computational overhead \cite{shen2024fastsam3d}, making them impractical for rapid, fully automated inference at deployment.
\par
To leverage the zero-shot capabilities of foundation models without inheriting their inference costs, recent works have explored the option of using them as external teachers to supervise task-specific student networks. Frameworks such as SemiSAM+ \cite{zhang2025semisam+} attempt to mitigate foundation-model noise by utilizing multi-prompt uncertainty within an online collaborative loop. However, because these methods still rely on dynamically querying the heavy foundation model during student training, they introduce severe computational and memory overheads. 
\\
In contrast, the proposed framework bypasses this computational bottleneck by entirely decoupling the foundation model from the inference phase. Rather than querying the heavy teacher dynamically, we extract its zero-shot knowledge offline and filter the resulting 3D pseudo-labels through a strict confidence threshold. These highly reliable targets are then distilled into a lightweight, task-specific 3D student network. At clinical deployment, the heavy foundation model is completely discarded. This allows the standalone student network to retain the high-accuracy benefits of the foundation model's priors while requiring zero interactive prompts and incurring minimal computational overhead.

\section{Methods}

\begin{figure}[tb!]
    \centering
    \includegraphics[width=1\linewidth]{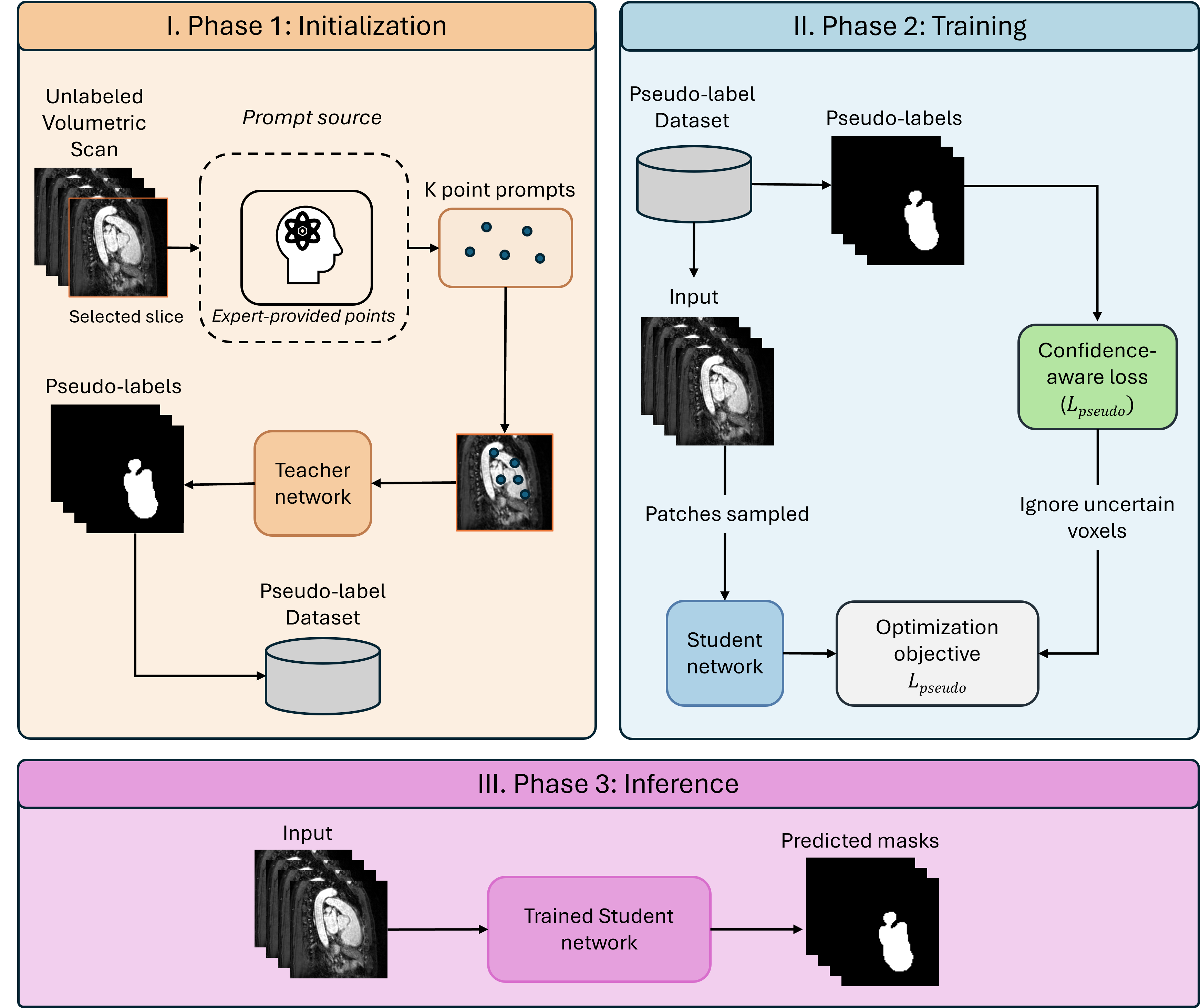}
    \caption{Overview of the proposed framework}
    \label{fig:overview}
\end{figure}
An overview of the proposed annotation-efficient segmentation framework for 3D medical segmentation is provided in \cref{fig:overview}.  The framework comprises three phases: (i) an initialization phase, in which a pretrained foundation model serves as a frozen teacher to generate offline volumetric pseudo-annotations; (ii) a training phase, in which a task-specific 3D student is optimized using confidence-aware pseudo-supervision; and (iii) an inference phase, in which the trained student independently produces volumetric segmentations. During the first phase, the framework receives as input only a sparse set of point prompts on a single 2D slice that can be provided by an expert clinician. The teacher utilizes these prompts to generate full-volume 3D pseudo-annotations, which serve as the sole training signal for the student network. The generated pseudo-annotations are stored as a static training dataset or pseudo-label dataset, and the teacher is not used during subsequent student optimization. Then, during the second phase, the student network is optimized exclusively on the pseudo-label dataset using a confidence-aware loss function that explicitly ignores uncertain voxels. As such, the teacher's role is strictly offline and is entirely omitted from the training loop. Finally, during the third phase, the trained student network is deployed independently for inference, generating predicted masks without any reliance on the teacher or further human interaction.

\subsection{Initialization Phase using Pseudo-label Generation}
Let $X=\{X_i\}_{i=1}^{N}$ denote a set of \textit{N} target-domain volumes, each volume $X_{i}$ is composed of 2D slices $\{s_{i,d}\}_{d=1}^{D_{i}}$, where $D_i$ denotes the total number of slices in the volume. Under the defined annotation-efficient segmentation setting, dense volumetric annotations are not considered during student training. Instead, the teacher is used to generate pseudo-annotations for each $s_i$. To generate the pseudo-annotations, the teacher requires a single selected 2D slice $s_i$ and a spatial prompt $P_i$ consisting of K positive points that reside within the target anatomy of that slice. Although any slice on which the target is visible can be selected, the middle foreground slice was selected in this study based on empirical ablations. In a clinical deployment scenario, an expert can provide the K point prompts directly via manual clicks on the chosen slice, eliminating the need to draw a dense 2D mask.
\par 
Then, the pretrained foundation model that is used as an offline teacher converts the sparse point prompt into full-volume supervisory information without task-specific fine-tuning. For each volume, the $K$-point prompt $P_i$ guides the teacher to generate a continuous foreground probability map. This probability map is then transformed into a 3D segmentation mask $\hat{y_i}$. In addition, confidence scores are computed from the teacher probability map and used to construct a reliability mask. Voxels with a lower confidence score than a defined threshold $\tau$ are assigned an “ignored” label that is used to exclude these voxels from the optimization process. The confidence scores for the remaining voxels are used as a weighting factor in the optimization process to ensure that the training is driven by the most reliable pseudo labels. The resulting pseudo-annotations and confidence scores form the supervision signals used to train the task-specific student network.

\subsection{Student Training with Confidence-Aware Pseudo Supervision}
\label{sec:training}
The proposed framework distills the point-prompted pseudo-annotations into a task-specific 3D student network. This distillation not only reduces computational overhead but also enables the student to learn continuous 3D spatial representations, effectively bridging the gap between independent 2D pseudo-annotations. Since the teacher performs zero-shot inference without target-domain fine-tuning, the generated pseudo-labels contain structural noise and boundary artifacts. To prevent the student from overfitting to the noisy pseudo-labels produced by the teacher, particularly near ambiguous anatomical boundaries, a hard masking strategy is employed. 
Let $S_\theta$ denote the task-specific 3D student network parameterized by $\theta$, and $S_\theta (X_i )$ represent its predicted full-volume probability map for an input volume $X_i$, where $i \in \{1,\ldots,N\}$ and \textit{N} denotes the total number of volumes in the dataset. Let $C_i$ represent the reliability mask containing the teacher's confidence scores $c_i (v)$ for voxel $v$, and $\hat{y_i}$ represent the corresponding discrete pseudo-labels. First, the voxel space is strictly partitioned based on the teacher’s predictive certainty. Voxels with confidence scores $c_i (v)$ falling below the threshold $\tau$ are mapped to a distinct “ignore” class and excluded from the optimization process. We define a binary validity mask $m_i (v) \in \{0,1\}$, where $m_i (v)=0$ for ignored uncertain voxels and $m_i (v)=1$ for confident pseudo-labels.
For the valid voxels that exceed the threshold ($m_i (v)=1$), the supervision signal is further modulated by a continuous confidence weight $w_i (v)$, defined as: 
\begin{equation}
  w_i (v)=w_{min}+(1-w_{min} )\cdot c_i (v)^\gamma
  \label{eq:cw}
\end{equation}

\noindent where $w_{min}$ denotes the baseline minimum weight assigned to all valid voxels and $\gamma$ is a scaling exponent that controls the non-linear curvature of the weighting function. This parameterization ensures that all valid pseudo-labels contribute a guaranteed minimum gradient signal, while disproportionately emphasizing highly confident teacher predictions over moderately confident ones to suppress residual uncertainty.
The network is supervised by a masked objective that restricts gradient updates strictly to the valid regions: 

\begin{equation}
    L_{pseudo}=\lambda_{ce}\cdot CE(w_i,S_\theta (X_i ),\hat{y_i};m_i )+\lambda_{dice}\cdot Dice(w_i,S_{\theta} (X_{i} ),\hat{y_{i}};m_{i} )
    \label{eq:loss}
\end{equation}

\noindent where $CE$ and $Dice$ are the masked Cross-Entropy and Dice Similarity Coefficient losses, respectively, and $\lambda_{ce}$ and $\lambda_{dice}$ are scalar balancing weights. By explicitly zeroing out the gradients in uncertain boundary regions, the network is not penalized by noisy teacher priors. Instead, it is free to naturally interpolate these boundaries based on the dense 3D anatomical context it learns from the confident regions.
At inference time, the optimized 3D student network alone is used to predict the final 3D volumes.

\par 
The complete training and inference procedure is summarized in \cref{alg:confidence_aware_training}. The process begins with an offline initialization phase, where the foundation model generates the 3D pseudo-annotations and corresponding continuous confidence maps. These confidence maps reflect the teacher model's predictive certainty and are utilized to weigh the optimization objective for each valid pseudo-label voxel. This pseudo-labeling procedure is conducted for the entire volume based on the sparse point prompts on the single selected slice. During the subsequent training phase, the student network is optimized over sampled 3D patches. For each forward pass, the confidence-aware pseudo-loss is computed exclusively over the valid voxels within the 3D patch, while gradients from uncertain regions are safely ignored.

\begin{algorithm}[t]
\caption{Confidence-Aware Pseudo-label Training \& Inference}
\label{alg:confidence_aware_training}

\textbf{Input:} Target 3D image volumes $X$, selected 2D slices $\{s_i\}_{i=1}^{N}$ with corresponding point prompts $P_i$, Teacher Model $F$, confidence threshold $\tau$.
\\
\underline{\textbf{Initialization Phase using Pseudo-label Generation:}}
\\
\For{each target volume $X_i$}{
    Prompt $F$ using $P_i$ on $s_i$.\\
    Generate hard pseudo-labels $\hat{y}_i$ and confidence maps $c_i$.\\
    Create the $m_i = [c_i \geq \tau]$ by assigning an ignore label where $m_i = 0$.\\
    Compute confidence weights $w_i$ for all valid voxels.\\
}

Initialize Student parameters $\theta_S$.\\

\underline{\textbf{Training Phase:}}\\

\Repeat{
    $L_{pseudo}$ is approx. stable \textbf{or} $(j \geq max\_iterations)$
}{
     Sample 3D patch $x_{patch}$ with augmentations.\\
    Compute $S(x_{patch}; \theta_S)$.\\
    Calculate $L_{pseudo}$ using $\hat{y_i}$, $m_i$ and $w_i$.\\
    Update $\theta_S$ using $\nabla{\theta_S}L_{pseudo}$.
}

\underline{\textbf{Inference Phase:}}\\

\For{each test volume}{
    Predict masks using optimized $\theta_S$.\\
}

\textbf{Output:} Final 3D segmentations and optimized student $\theta_S$.\\

\end{algorithm}

\section{Experimental Setup}
\subsection{Datasets}
 Left-atrial segmentation involves ambiguous anatomical boundaries and small or thin structures, while dense volumetric annotation is labor-intensive. These characteristics have motivated its extensive use for evaluating semi-supervised segmentation methods under limited annotation \cite{wang2025synergy,wu2022mutual,xiong2021global}. As such, experiments were conducted on the training sets of two benchmark 3D cardiac MRI datasets for left-atrial segmentation, namely the Medical Segmentation Decathlon (MSD) Task02 Heart \cite{antonelli2022medical} and the 2018 Left Atrium Segmentation Challenge (LA) dataset \cite{xiong2021global}. The MSD Heart dataset consists of 20 labeled 3D mono-modal MRI volumes with expert annotations of the left atrium and is characterized by a small training dataset with large variability \cite{antonelli2022medical}. The LA dataset provides 100 labeled 3D late-gadolinium-enhanced MRI (LGE-MRI) volumes from patients with atrial fibrillation, where segmentation is challenged by variable image intensities associated with contrast enhancement. In this context, MSD Heart provides a small-data setting, while LA provides a larger-scale cardiac MRI setting for evaluating whether the same point-prompted pseudo-labeling strategy remains effective when more training volumes are available. Both datasets are formulated as binary segmentation tasks, where the left atrium is treated as foreground and all remaining voxels as background.

\subsection{Implementation Details}
MedSAM2 is used as the frozen foundation-model teacher \cite{ma2025medsam2}. To generate the initial pseudo-annotations, MedSAM2 is prompted using five positive points sampled from the middle foreground slice of each volume. This specific configuration is derived from an ablation study we performed using different prompts, which demonstrate that middle-slice interaction with a five-point budget maximizes pseudo-annotation quality while maintaining strict annotation efficiency. The task-specific student network is instantiated as a 3D nnU-Net and optimized using Stochastic Gradient Descent (SGD) with an initial learning rate of $0.01$, momentum set to $0.99$ and weight decay $3\times10^{-5}$. The student is trained for a total of 25,000 iterations  with batch size 4. Prior to optimization, the offline pseudo-annotations generated by the teacher are formulated into a static training dataset. Regarding voxel-wise reliability, the optimal value for $\tau$ is determined with an  ablation study, yielding $ \tau=0.5$. For the valid pseudo-labeled regions, soft weighting is applied by setting $w_{min}=0.7$ and $\gamma=1.5$. The pseudo-supervision loss components are balanced equally, with $\lambda_{ce}=0.5$, $\lambda_{dice}=0.5$, following standard nnU-Net training conventions.

\subsection{Evaluation Protocol}
Segmentation performance is evaluated using region-overlap, boundary-distance, and classification-oriented segmentation metrics. Region overlap is measured using the Dice similarity coefficient (DSC) and intersection-over-union (IoU). Boundary accuracy is assessed using the 95th percentile Hausdorff distance (HD95) and average symmetric surface distance (ASSD), which capture surface-level deviations between the predicted and reference segmentations \cite{taha2015metrics}. Results are reported as mean across 5 cross-validation folds.
\par 
The proposed framework was compared with four state-of-the-art  semi-super-\newline vised segmentation methods, \textit{i.e.}, Mean Teacher \cite{tarvainen2017mean}, MC-Net+ \cite{wu2022mutual}, SGRS-Net \cite{wang2025synergy}, and AD-MT \cite{zhao2024alternate}. Since these methods were originally designed for settings containing fully annotated and unannotated volumes, their supervision interface is adapted to the considered slice-sparse setting. For every training volume, only the ground-truth mask of the selected middle foreground slice is provided to the supervised loss. All remaining voxels are treated as unannotated and are processed using the original semi-supervised objective of each method. Therefore, each of these methods receives one annotated slice per training volume, whereas the proposed framework uses only five positive points sampled from the same slice. This comparison is intended to assess performance under sparse human supervision, where conventional semi-supervised approaches rely solely on task-specific supervision without leveraging external foundation-model priors. The performance of MedSAM2 is also included to quantify the effect of student distillation and a fully supervised 3D nnU-Net (FS) trained with complete volumetric annotations is reported as the performance upper bound.

\section{Results}
\label{sec:results}

\subsection{Ablation Studies}
Ablation experiments were conducted to determine the optimal prompt configuration, the sensitivity of student training to the confidence threshold $\tau$, and the contributions of the proposed training mechanisms. 

\subsubsection{Prompt configuration}
To determine the optimal prompting configuration, we varied the axial slice used to initialize MedSAM2 and the number of positive points sampled from its foreground mask. For each volume, middle denotes the central slice of the foreground extent, largest the slice with the greatest foreground area, random a randomly selected foreground-containing slice, and boundary low and boundary high the lowest- and highest-indexed foreground-containing slices, respectively. From each selected slice, 1, 5, or 10 positive points were sampled within the foreground region and provided to MedSAM2. 

\renewcommand{\tabularxcolumn}[1]{m{#1}}
\newcolumntype{Y}{>{\centering\arraybackslash}X}

\begin{table}[tb]
  \caption{Effect of the slice selection strategy and number of positive point prompts on MedSAM2 pseudo-annotation quality, averaged across MSD Heart and LA.
  }
  \label{tab:ablstrat}
  \centering
   \setlength{\tabcolsep}{4pt}
  \begin{tabularx}{\linewidth}{@{}cYYYcc@{}}
    \toprule
    Slice Strategy & No. points & DSC (\%) $\uparrow$ & IoU (\%) $\uparrow$ & HD95 (mm) $\downarrow$	& ASSD (mm) $\downarrow$ \\
    \midrule
    Middle	& 1 &	86.56	& 77.03 &	10.66 &	2.60 \\
    Middle &	5 &	87.43 &	77.86 &	9.43 & 	2.05 \\
    Middle &	10 &	87.26 &	77.60 &	9.66 &	2.09 \\
    Largest &	1 &	82.33 &	72.19  &	13.91 &	3.83 \\
    Largest &	5 &	86.25 &	76.13 &	10.71 &	2.27 \\
    Largest &	10 &	86.09 &	75.89 &	10.51 &	2.28 \\
    Random &	1 &	72.22 &	61.92 &	24.02 &	7.66 \\
    Random &	5 &	75.96 &	65.43 &	20.88 &	6.06 \\
    Random &	10 &	75.86 &	65.14 &	21.10 &	6.03 \\
    Boundary low &	1 &	72.64 &	59.46 &	27.60 &	6.45 \\
    Boundary low &	5 &	74.29 &	60.84 &	26.24 &	5.79 \\
    Boundary low &	10 &	74.49 &	61.18 &	25.91 &	5.73 \\
    Boundary high &	1 &	51.93 &	38.89 &	40.67 &	12.50 \\
    Boundary high &	5 &	57.23 &	43.35 &	36.31 &	10.12 \\
    Boundary high &	10 &	57.86 &	43.76 &	35.76 &	9.68 \\
  \bottomrule
  \end{tabularx}
  \vspace{0.4em}
    \parbox{\linewidth}{\footnotesize
    \textit{Note:} $\uparrow$ indicates that higher values are better;
    $\downarrow$ indicates that lower values are better.}
\end{table}

As detailed in \cref{tab:ablstrat}, the selection of the prompt slice significantly impacts the quality of the foundation model's generated pseudo-annotations. The middle fore-ground slice consistently yields the highest segmentation performance, achieving a peak Dice Similarity Coefficient (DSC) of 87.43\% and minimizing boundary errors (2.05 ASSD). While prompting the largest foreground slice produces reasonable but slightly inferior results, performance degrades drastically when relying on random sampling or boundary slices. Notably, prompting the upper boundary fails to capture the core anatomical structure entirely, resulting in a severe performance drop (57.86\% DSC). Within the optimal middle-slice configuration, we evaluated prompts containing 1, 5, and 10 positive points. Although a single point provides a remarkably strong initialization (86.56\% DSC), it often proves insufficient to fully delineate complex anatomical boundaries, resulting in higher surface distances (2.60 ASSD). Increasing the number of positive points to 5 points yields a clear improvement in both region overlap and boundary precision. Importantly, further increasing the budget to 10 points provides no additional quantitative benefit, yielding essentially tied metrics. Consequently, the middle-slice, 5-point prompting strategy was selected as the default configuration for all main experiments, as it successfully maximizes pseudo-annotation quality while strictly minimizing the simulated annotation bur-den.

\subsubsection{Confidence threshold}
The confidence threshold $\tau$ controls the trade-off between retaining useful pseudo-supervision and excluding uncertain predictions.  \cref{tab:ablconf} examines the sensitivity of student training to this threshold. For each row, the same value of $\tau$ is applied to both datasets before their performance is averaged. 
As can be observed, the framework remains stable across the evaluated confidence thresholds. Specifically, a confidence threshold of $ \tau=0.5$ yields the strongest overall cross-dataset performance, achieving an average DSC of 89.05\% and an average IoU of 80.38\%, together with highly competitive surface errors. Given this robust and consistent behavior across varying data distributions, $\tau=0.5$ is established as the default configuration for all remaining experiments.

\begin{table}[tb]
  \caption{ Effect of the confidence threshold $\tau$ on student segmentation performance, averaged across MSD Heart and LA.
  }
  \label{tab:ablconf}
  \centering
  \setlength{\tabcolsep}{15pt}
    
  \begin{tabularx}{\linewidth}{@{}Ycccc@{}}
  
    \toprule
    $\tau$	& DSC (\%) $\uparrow$ & IoU (\%) $\uparrow$ & HD95 (mm) $\downarrow$	& ASSD (mm) $\downarrow$ \\
    \midrule
    0.5 &	89.05 &	80.38 &	8.04 &	1.77 \\
    0.6 &	88.76 &	79.92 &	8.51 &	1.84 \\
    0.7 &	88.91 &	80.17 &	8.33 &	1.86 \\
    0.8 &	88.42 &	79.52 &	8.46 &	1.89 \\
  \bottomrule
  \end{tabularx}
  \vspace{0.4em}
    \parbox{\linewidth}{\footnotesize
    \textit{Note:} $\uparrow$ indicates that higher values are better;
    $\downarrow$ indicates that lower values are better.}
\end{table}

\subsubsection{Training mechanisms}
The contributions of ignore masking (IM) and confidence weighting (CW) are assessed by comparing four student-training configurations. Student PL is trained directly on the MedSAM2 pseudo-labels without either mechanism. Student IM applies the ignore mask $m_i (v)$ without confidence weighting, whereas Student CW applies confidence weighting without excluding uncertain voxels. Lastly, the proposed, Student IM+CW, configuration combines IM and CW. As observed in  \cref{tab:ablmech}, combining IM and CW provides the highest mean overlap and the lowest mean surface distances, reaching 89.05\% DSC, 80.38\% IoU, 8.04 HD95, and 1.77 ASSD. Removing either mechanism reduces the mean performance, while IM without CW produces the lowest mean results among the evaluated configurations.

\begin{table}[tb]
  \caption{ Effect of the proposed training mechanisms on segmentation performance, averaged across MSD Heart and LA.
  }
  \label{tab:ablmech}
  \centering
  \setlength{\tabcolsep}{6pt}
    
  \begin{tabularx}{\linewidth}{@{}Ycccc@{}}
    \toprule
    Training mechanism & DSC (\%) $\uparrow$ & IoU (\%) $\uparrow$ & HD95 (mm) $\downarrow$	& ASSD (mm) $\downarrow$ \\
    \midrule
    Student PL &	88.78 &	79.97 &	8.87 &	2.01 \\
    Student IM &	87.74 &	78.48 &	9.86 &	2.21 \\
    Student CW &	88.57 &	79.64 &	9.11 &	2.02 \\
    Student IM+CW &	89.05 &	80.38 &	8.04 &	1.77 \\
  \bottomrule
  \end{tabularx}
  \vspace{0.4em}
    \parbox{\linewidth}{\footnotesize
    \textit{Note:} $\uparrow$ indicates that higher values are better;
    $\downarrow$ indicates that lower values are better.}
\end{table}

\subsection{Quantitative Results}

\cref{tab:quant} presents the quantitative comparison on MSD Heart and LA. Under the considered slice-sparse protocol, the prompt-based approaches consistently outperform the adapted SOTA semi-supervised methods on both datasets. Mean Teacher and MC-Net+ provide the strongest results among the semi-supervised methods, with MC-Net+ attaining the highest performance within this group across all metrics. Nevertheless, both perform worse compared to MedSAM2, while the proposed framework further improves MedSAM2 and achieves the best mean performance among the non-fully supervised methods on both datasets.

\begin{table}[tb!]
    \centering
    \caption{Quantitative comparison with state-of-the-art semi-supervised
    and prompt-based methods on MSD Heart and LA. A fully supervised 3D
    nnU-Net is included as an upper bound. Bold indicates the best performance
    among the non-fully supervised methods.}
    \label{tab:quant}

    \small
    \setlength{\tabcolsep}{4pt}

    \begin{tabularx}{\linewidth}{
        @{}
        Ycc
        cccc
        @{}
    }
        \toprule
        Dataset
        & Method
        & Type
        & \shortstack{DSC \\ (\%) $\uparrow$}
        & \shortstack{IoU \\ (\%) $\uparrow$}
        & \shortstack{HD95 \\ (mm) $\downarrow$}
        & \shortstack{ASSD \\ (mm) $\downarrow$} \\
        \midrule

        \multirow{7}{*}{\shortstack{MSD\\Heart}}
        & FS
        & Fully supervised
        & 93.28 & 87.47 & 2.96 & 0.77 \\
        
        & Mean Teacher~\cite{tarvainen2017mean}
        & Semi-supervised
        & 83.52 & 72.30 & 14.51 & 3.05 \\

        & MC-Net+~\cite{wu2022mutual}
        & Semi-supervised
        & 84.42 & 73.54 & 13.81 & 2.79 \\

        & SGRS-Net~\cite{wang2025synergy}
        & Semi-supervised
        & 79.58 & 66.35 & 15.15 & 3.42 \\

        & AD-MT~\cite{zhao2024alternate}
        & Semi-supervised
        & 80.18 & 67.80 & 17.43 & 3.77 \\

        & MedSAM2~\cite{ma2025medsam2}
        & Prompt-based
        & 86.74 & 76.77 & 8.90 & 1.93 \\
        & Proposed
        & Prompt-based
        & \textbf{88.63}
        & \textbf{79.65}
        & \textbf{8.13}
        & \textbf{1.70} \\

        \midrule

        \multirow{7}{*}{LA}
        & FS
        & Fully supervised
        & 92.08 & 85.46 & 4.54 & 1.21 \\

        & Mean Teacher~\cite{tarvainen2017mean}
        & Semi-supervised
        & 60.61 & 45.93 & 96.66 & 30.76 \\

        & MC-Net+~\cite{wu2022mutual}
        & Semi-supervised
        & 62.29 & 46.93 & 57.79 & 16.09 \\

        & SGRS-Net~\cite{wang2025synergy}
        & Semi-supervised
        & 29.65 & 19.19 & 182.13 & 77.51 \\

        & AD-MT~\cite{zhao2024alternate}
        & Semi-supervised
        & 49.13 & 35.44 & 127.78 & 47.87 \\

        & MedSAM2~\cite{ma2025medsam2}
        & Prompt-based
        & 88.12 & 78.95 & 9.96 & 2.17 \\
        \vspace{0.4em}
        & Proposed
        & Prompt-based
        & \textbf{89.47}
        & \textbf{81.11}
        & \textbf{7.95}
        & \textbf{1.85} \\

        \bottomrule
    \end{tabularx}

    \vspace{0.4em}

    \parbox{\linewidth}{\footnotesize
        \textit{Note:} $\uparrow$ indicates that higher values are better;
        $\downarrow$ indicates that lower values are better.
    }
\end{table}

\subsubsection{MSD Heart Dataset}
On the smaller MSD Heart dataset, the proposed framework achieves 88.63\% DSC, 79.65\% IoU, 8.13 mm HD95, and 1.70 mm ASSD. MC-Net+ and Mean Teacher are the strongest semi-supervised methods, reaching 84.42\% and 83.52\% DSC, respectively. Compared with MC-Net+, the proposed framework provides relative improvements of 5.00\% in DSC and 8.30\% in IoU, while reducing HD95 by 41.10\% and ASSD by 39.10\%. Compared with MedSAM2, student distillation provides relative improvements of 2.20\% in DSC and 3.80\% in IoU, together with reductions of 8.70\% in HD95 and 11.90\% in ASSD. Relative to MedSAM2, the strongest competing non-fully supervised method, the proposed framework reduces the DSC gap to the fully supervised upper bound by 28.90\%, while requiring only five positive point prompts per volume.

\subsubsection{LA Dataset}
On the larger LA dataset, the proposed framework again achieves the best mean performance among the non-fully supervised methods, reaching 89.47\% DSC, 81.11\% IoU, 7.95 mm HD95, and 1.85 mm ASSD. MC-Net+ and Mean Teacher remain the strongest semi-supervised methods, but reach only 62.29\% and 60.61\% DSC, respectively, indicating that their performance degrades substantially under the slice-sparse adaptation. Compared with MC-Net+, the proposed framework provides relative improvements of 43.60\% in DSC and 72.80\% in IoU, while reducing HD95 by 86.20\% and ASSD by 88.50\%. Compared with MedSAM2, it provides relative im-provements of 1.50\% in DSC and 2.70\% in IoU, together with reductions of 20.20\% in HD95 and 14.70\% in ASSD. Relative to MedSAM2, the proposed framework reduces the DSC gap to the fully supervised upper bound by 34.10\%.

\subsection{Qualitative Results}

\begin{figure}[t]
    \centering

    \begingroup
    \setlength{\tabcolsep}{0pt}

    \begin{tabularx}{\linewidth}{
        @{}*{6}{>{\centering\arraybackslash}X}@{}
    }
        \scriptsize
        \makebox[0pt][c]{\hspace*{-8.0mm}GT} &
        \makebox[0pt][c]{\hspace*{-6mm}FS} &
        \makebox[0pt][c]{\hspace*{-1.5mm}Mean Teacher} &
        \makebox[0pt][c]{\hspace*{4mm}MC-Net+} &
        \makebox[0pt][c]{\hspace*{5mm}MedSAM2} &
        \makebox[0pt][c]{\hspace*{10mm}Proposed}
        \\[2pt]

        \multicolumn{6}{@{}c@{}}{%
            \includegraphics[width=\linewidth]{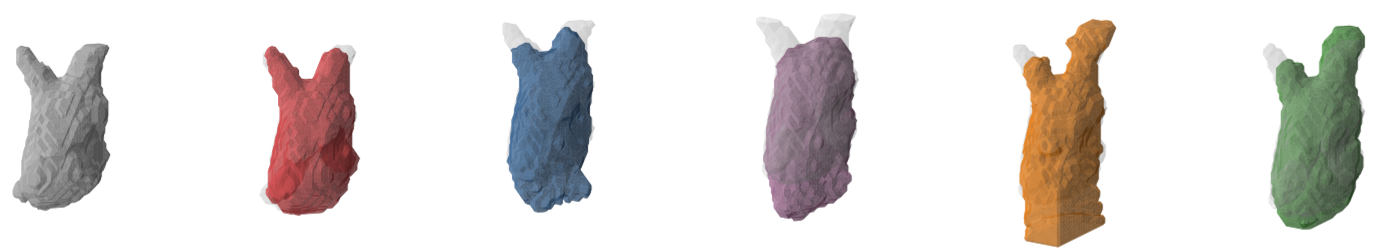}
        }
        \\[-1pt]

        \multicolumn{6}{@{}c@{}}{\small (a)}
        \\[4pt]

        \multicolumn{6}{@{}c@{}}{%
            \includegraphics[width=\linewidth]{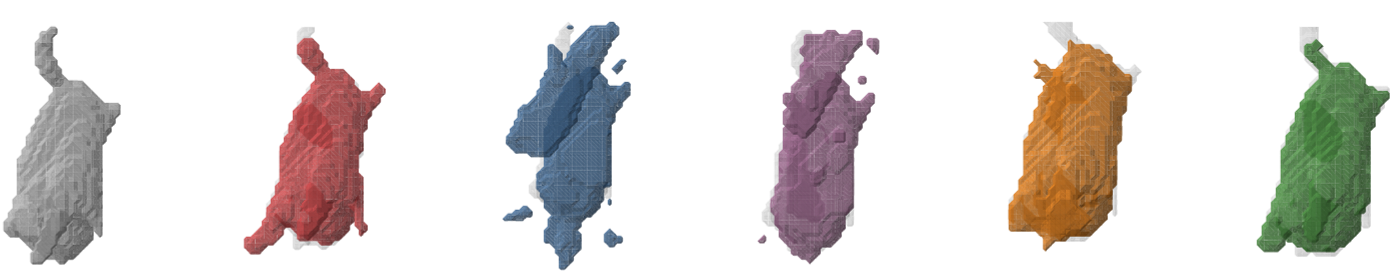}
        }
        \\[-1pt]

        \multicolumn{6}{@{}c@{}}{\small (b)}
    \end{tabularx}

    \endgroup

    \caption{Qualitative comparison of 3D reconstructions on representative
    held-out volumes from (a) MSD Heart and (b) LA. GT denotes the ground
    truth, and FS denotes the fully supervised 3D nnU-Net.}
    \label{fig:qualitative_comparison}
\end{figure}

\cref{fig:qualitative_comparison} presents representative qualitative 3D reconstructions for two representative held-out volumes: one from MSD Heart (a) and one from LA (b). For each case, the ground truth (GT) is compared with FS, Mean Teacher, MC-Net+, MedSAM2, and the proposed framework. The reconstructions were generated by stacking the binary segmentation masks from all axial slices in their original order to form a complete 3D volume and subsequently rendering its foreground surface. Across both datasets, the proposed framework more closely preserves the ground-truth morphology and surface continuity than the adapted semi-supervised methods and the direct MedSAM2 pre-dictions, while approaching the fully supervised segmentation quality.

\noindent In the MSD Heart case (\cref{fig:qualitative_comparison}a), Mean Teacher introduces small, disconnected com-ponents, while MC-Net+ and MedSAM2 produce enlarged, irregular reconstructions. The proposed framework better preserves the narrow main body and superior bifurca-tion, although some fine distal structures remain incomplete. In the LA case (\cref{fig:qualitative_comparison}b), the semi-supervised methods exhibit substantial over-segmentation and disconnect-ed regions, whereas MedSAM2 produces geometrically coarse and locally irregular boundaries. The proposed reconstruction suppresses these artifacts and more closely follows the compact, continuous morphology of the GT and FS reference, consistent with the quantitative improvements in surface-distance metrics.

\section{Conclusion}

This study introduced an annotation-efficient, confidence-aware teacher–student framework for 3D medical image segmentation under extreme annotation sparsity. Using only five positive points placed on a single slice of each training volume, the framework employs an offline foundation model to generate volumetric pseudo-annotations for training a task-specific 3D student. To reduce the influence of unreliable pseudo-labels, the proposed objective combines ignore masking, which excludes low-confidence voxels, with confidence weighting, which modulates the contribution of the retained supervision. This design limits the propagation of uncertain teacher predictions, particularly near ambiguous anatomical boundaries. Once trained, the student operates independently of MedSAM2 and does not require interactive prompts, enabling fully automated 3D inference without the computational overhead of running the foundation model at deployment.
\par
Experiments on the MSD Heart and LA datasets demonstrated that the proposed framework consistently outperformed the prompt-based MedSAM2 and the adapted semi-supervised methods in both region-overlap and surface-distance metrics. The comparison also revealed different levels of robustness under the same slice-sparse supervision regime. While the semi-supervised methods retained moderate segmentation performance on MSD Heart, their accuracy deteriorated markedly on the larger LA dataset. In contrast, the proposed framework maintained robust performance across both datasets, achieving DSC values of 88.63\% and 89.47\%, and achieved up to a 43.60\% relative improvement in DSC compared with the best-performing semi-supervised method. Accordingly, this comparison also illustrates the benefit of incorporating external foundation-model knowledge under extreme sparse supervision, rather than isolating the contribution of the student optimization alone. In addition, the proposed framework reduced HD95 by up to 20.20\% and ASSD by up to 14.70\% compared to MedSAM2, while accounting for up to 34.10\% of the DSC gap between MedSAM2 and the fully supervised network. Despite these merits, performance may be affected when sparse prompting fails to capture fine or disconnected anatomical structures, or when erroneous teacher predictions are assigned high confidence. 
\par 
Consistent performance across two datasets of different scales provides encouraging evidence of robustness within left-atrial MRI, while broader evaluation across other anatomies and modalities would further assess generalizability. Additionally, the empirically selected confidence threshold provides a common operating point across both datasets, while adaptive confidence estimation could be explored as an alternative for adjusting pseudo-supervision during training. Future work will thus explore multi-class anatomical segmentation and investigate self-supervised learning strategies to further improve segmentation performance. Adaptive or learned confidence selection, together with improved teacher-confidence calibration and complementary uncertainty estimation, will be explored to better identify unreliable pseudo-labels and account for dataset- or sample-specific uncertainty. The feasibility of online teacher–student variants will also be investigated, with particular attention to computational efficiency and the trade-off between segmentation accuracy and training cost. Broader experimental validation will also consider diverse anatomies, imaging modalities, and pathologies, together with comparisons against additional knowledge-distillation and pseudo-label-refinement approaches.

\section*{Acknowledgements}
%Please insert your acknowledgments here.
This work is part of the European project SEARCH (https://ihi-search.eu/), which is supported by the Innovative Health Initiative Joint Undertaking (IHI JU) under grant agreement No. 101172997. The JU receives support from the European Union’s Horizon Europe research and innovation programme and COCIR, EFPIA, Europa Bio, MedTech Europe, Vaccines Europe, Medical Values GmbH, Corsano Health BV, Syntheticus AG, Maggioli SpA, Motilent Ltd, Ubitech Ltd, Hemex Benelux, Hellenic Healthcare Group, German Oncology Center, Byte Solutions Unlimited, AdaptIT GmbH. Views and opinions expressed are, however, those of the author(s) only and do not necessarily reflect those of the aforementioned parties. Neither of the aforementioned parties can be held responsible for them. 
\bibliographystyle{splncs04}
\bibliography{main}
\end{document}